\pdfoutput=1

\documentclass[11pt]{article}

\usepackage[preprint]{acl}

\usepackage{times}
\usepackage{latexsym}
\usepackage{booktabs}
\usepackage{multirow}
\usepackage{multicol}
\usepackage{graphicx}
\usepackage{makecell}
\usepackage{amssymb}
\usepackage{amsmath}  
\usepackage[T1]{fontenc}

\usepackage[utf8]{inputenc}

\usepackage{microtype}

\usepackage{inconsolata}

\usepackage{graphicx}

\title{MMLU-CF: A Contamination-free Multi-task Language Understanding Benchmark}

\author{\textbf{Qihao Zhao}\;\;\;\textbf{Yangyu Huang}\thanks{Corresponding author}\;\;\;\;\textbf{Tengchao Lv}\;\;\;\textbf{Lei Cui}\;\;\;\textbf{Furu Wei}\;\;\;\\
\textbf{Qinzheng Sun}\;\;\;\textbf{Ying Xin}\;\;\;\textbf{Shaoguang Mao}\;\;\;\textbf{Xin Zhang}\;\;\;\textbf{Qiufeng Yin}\;\;\;\textbf{Scarlett Li} 
\\
Microsoft Research}

\begin{document}

\maketitle

\begin{abstract}
Multiple-choice question (MCQ) datasets like Massive Multitask Language Understanding (MMLU) are widely used to evaluate the commonsense, understanding, and problem-solving abilities of large language models (LLMs).
However, the open-source nature of these benchmarks and the broad sources of training data for LLMs have inevitably led to benchmark contamination, resulting in unreliable evaluation results.
To alleviate this issue, we propose a contamination-free and more challenging MCQ benchmark called MMLU-CF.
This benchmark reassesses LLMs' understanding of world knowledge by averting both unintentional and malicious data leakage.
To avoid unintentional data leakage, we source data from a broader domain and design three decontamination rules.
To prevent malicious data leakage, we divide the benchmark into validation and test sets with similar difficulty and subject distributions.
The test set remains closed-source to ensure reliable results, while the validation set is publicly available to promote transparency and facilitate independent verification.
Our evaluation of mainstream LLMs reveals that the powerful GPT-4o achieves merely a 5-shot score of 73.4\% and a 0-shot score of 71.9\% on the test set, which indicates the effectiveness of our approach in creating a more rigorous and contamination-free evaluation standard.
\end{abstract}

\begin{figure}[!ht]
\centering
\includegraphics[width=1\columnwidth]{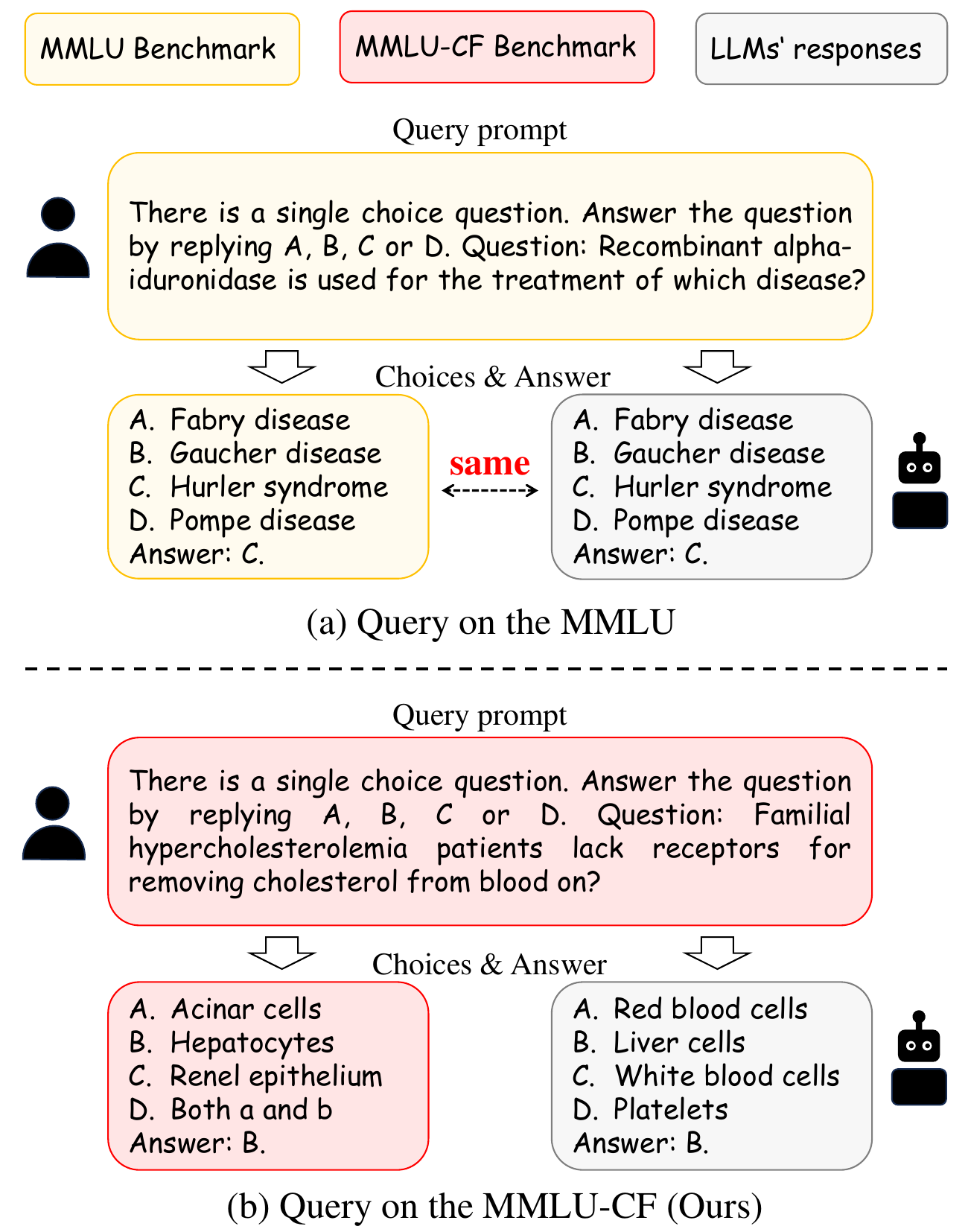}
\caption{(a) An instance of leakage in MMLU.
When questions are used as prompt from the MMLU, certain LLMs, due to their memorization capabilities, directly provide \textbf{choices identical to the original ones}.
(b) When questions are used as prompt from the MMLU-CF, LLMs only provide guessed choices.
This indicates that the MMLU test set suffers from data contamination and memorization by some LLMs, while the proposed MMLU-CF avoids such leakage.}
\label{fig_1}
\end{figure}

\section{Introduction}

\begin{figure*}[t]
\centering
\includegraphics[width=2\columnwidth]{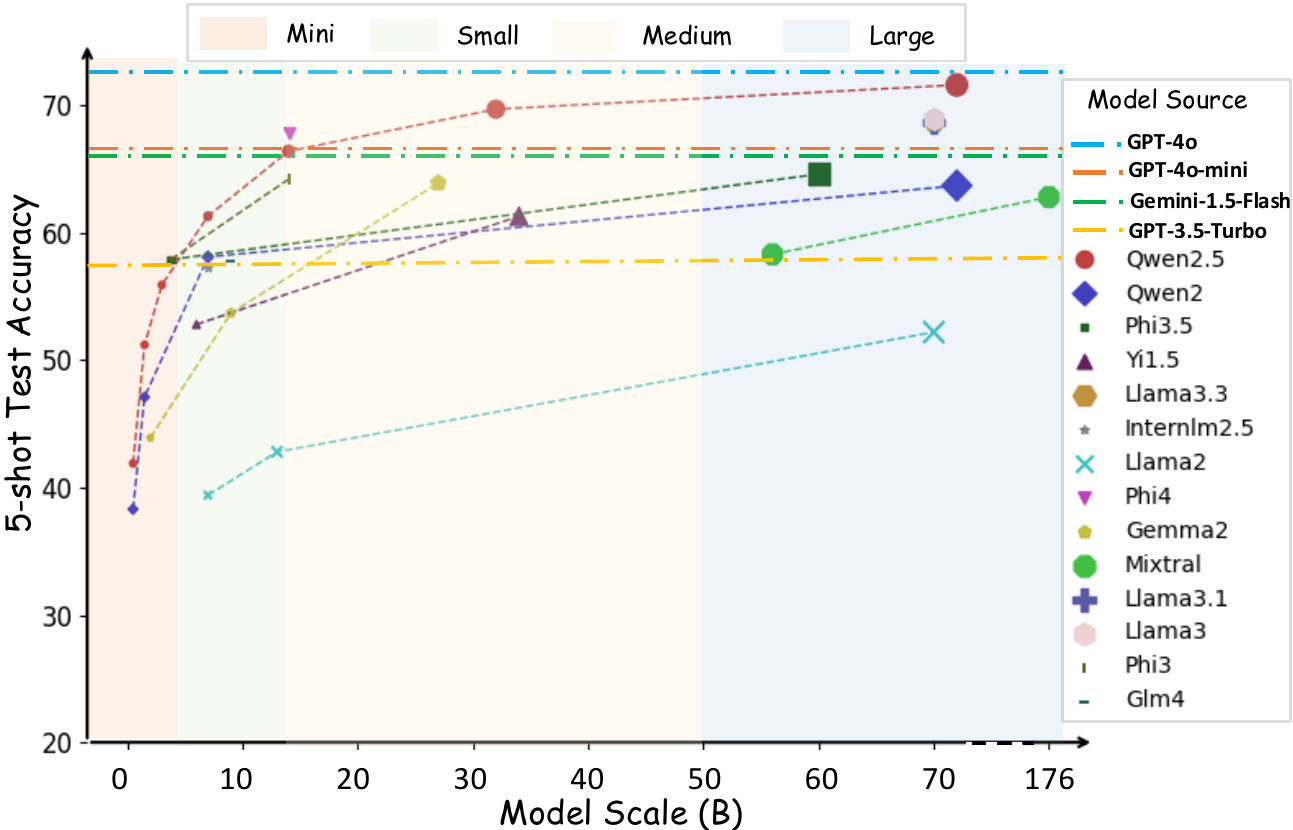}
\caption{
The 5-shot results on the MMLU-CF test set encompass mainstream open-source models ranging from 0.5 billion (B) to 176 billion (B) parameters, and including closed-source API models.
}
\label{fig_2}
\end{figure*}

Given the emergence of powerful capabilities in large language models (LLMs) such as GPT-4 \cite{achiam2023gpt}, Llama \cite{meta2024llama3}, Gemini \cite{reid2024gemini}, and Claude-3 \cite{anthropic2023claude3}, the evaluation of these models have become particularly important for understanding their strengths and limitations.
Consequently, a number of benchmarks covering reasoning \cite{hendrycks2020mmlu, wang2024mmlupro}, reading comprehension, mathematics \cite{cobbe2021training}, science \cite{rein2023gpqa}, and coding \cite{yu2023wavecoder} have been explored and released.
Among them, Massive Multitask Language Understanding (MMLU) \cite{hendrycks2020mmlu} is a widely used multiple-choice question (MCQ) gold standard benchmark because it covers various disciplines and difficulty levels, allowing for a comprehensive evaluation of LMMs' performance across diverse domains. 

However, data leakage or contamination, where LLMs inadvertently encounter benchmark data during training, can compromise the effectiveness, reliability, and fairness of these evaluations \cite{deng2024investigating, roberts2023cutoff}, termed general contamination.
Additionally, due to the public availability of benchmarks and the ability of LLMs to memorize data \cite{carlini2023quantifying}, instances of malicious contamination may occur.
As illustrated in Figure \ref{fig_1}, we observe that when only given the questions, some LLMs directly provide the choices and answers, where the choices are exactly the same as those in the MMLU test set.
This indicates that the benchmark may have been maliciously added to the training set and that LLMs have memory for these questions. 

To fairly investigate the world knowledge of LLMs, we propose \textit{MMLU-CF}, a contamination-free multiple-choice question benchmark for LLMs. To minimize the risk of benchmark exposure and contamination, we perform five key processing steps for the data: (1) MCQ Collection, (2) MCQ Cleaning, (3) Difficulty Sampling, (4) LLMs Checking, (5) Contamination-Free Processing.
In the contamination-free processing step, we employed three rules to rewrite the questions.
For detailed information, please refer to Section \ref{sec:dataset_construction_pipeline}.
For humans, rewriting the questions without changing their meaning does not affect their judgment.
However, if the model has seen the question and only memorizes it, the rewriting will affect the model’s judgment of the question.
Finally, we construct the MMLU-CF consisting of 10,000 questions for the test set and another 10,000 questions for the validation set.
To prevent malicious exposure, the test set remains closed-source \cite{zhang2024GSM1K}, while the validation set is open-source for evaluation.

We benchmark leading open-source and closed-source LLMs on the MMLU-CF test and validation sets, including GPT-4o \cite{openai2024gpt4o}, GPT-4o-mini \cite{gpt4omini}, Gemini \cite{reid2024gemini}, Qwen \cite{bai2023qwen, qwen2.5}, Llama \cite{meta2024llama3}, Phi \cite{abdin2024phi, abdin2024phi4technicalreport}, and many more. The 5-shot test results are briefly summarized in Figure \ref{fig_2}. Closed-source \textbf{API models} such as GPT-4o perform consistently well on the MMLU-CF 5-shot test, with GPT-4o leading at 73.4\%. This result is significantly lower than the 88.0\% on MMLU, highlighting the challenging and contamination-free nature of MMLU-CF. GPT-4o-mini, despite being lightly designed, achieves 65.5\% accuracy. Among \textbf{large models} (>50B parameters), Qwen2.5-72B-instruct stands out with a strong 5-shot test score of 71.6\%, approaching GPT-4-level performance. Llama-3.3-70B-instruct also achieves impressive results with a test score of 68.8\%, while other models such as Qwen1.5-72B-chat and Llama-2-70B-chat perform lower at 59.8\% and 52.2\%, respectively. For \textbf{medium models},  Qwen2.5-32B-instruct performs strongly with a test score of 69.7\%. Phi-4-14B also achieves an impressive 67.8\% on the 5-shot test, outperforming several larger models, such as Qwen2-72B (63.7\%) and Mixtral-8x22B (62.8\%), demonstrating the efficiency of its architecture. For \textbf{small models} Qwen2.5-7B-instruct delivers notable results at 61.3\%, outperforming other models in this category, such as Glm-4-9B-chat (57.8\%) and Llama-3-8B-instruct (57.3\%). For \textbf{mini models}, Phi-3.5/3-mini-instruct achieves a 5-shot test score of 57.9\%, leading the segment. Qwen2.5-3B-instruct performs slightly lower at 55.9\% but still outperforms other models in its class, demonstrating the strength of its design.

Additionally, the performances of mainstream models on the test set and validation set are quite approaching. In the future, we will publicly release the validation set to facilitate independent verification. If the accuracy gap between the test set and validation set increases, it indicates that the validation set is gradually becoming contaminated. However, we still have an uncontaminated test set to reliably evaluate the performance of LLMs.

\section{Related Work}
\subsection{The Benchmark of LLMs} 

In the field of natural language processing (NLP), benchmarks play a crucial role in evaluating and comparing the performance of different large language models \cite{wang2018glue, cobbe2021training, hendrycks2020mmlu, wang2024mmlupro, IFEval, humaneval, rein2023gpqa, zhang2024xlam, math}. They serve as a common ground for fair comparison, fostering transparency and reproducibility in research. For instance, GLUE \cite{wang2018glue, sarlin2020superglue} is a collection of nine different tasks designed to evaluate the natural language understanding capabilities of models. 
GSM8K \cite{cobbe2021training} is a benchmark dataset of 8,000 high-quality, linguistically diverse grade school math word problems. It is designed to evaluate the problem-solving abilities of language models, requiring a combination of language understanding and mathematical reasoning. MMLU \cite{hendrycks2020mmlu} is a benchmark designed to evaluate a model’s multitask learning capabilities across a diverse set of 57 tasks, including high school mathematics, college-level biology, law, and more, focusing on testing the model’s generalization ability across different domains. Building upon this, MMLU-Pro \cite{wang2024mmlupro} enhances the benchmark by introducing more challenging, reasoning-focused questions and expanding the choice set from four to ten choices, shifting the emphasis from knowledge retrieval to reasoning. Further, MMLU-Pro+ \cite{asgari2024mmluproplus} extends MMLU-Pro by assessing shortcut learning and higher-order reasoning in large language models, offering a comprehensive evaluation of both reasoning depth and model robustness.

These benchmarks have become standard tools in the evaluation of large language models due to their widespread adoption and comprehensive coverage of various domains. However, these benchmarks, such as MMLU, MMLU-Pro, and MMLU-Pro+, focus on the breadth, reasoning, and difficulty of the questions without considering contamination prevention.

\subsection{The Contamination-free Benchmark}
Several benchmark datasets have been introduced for contamination-free evaluation. LatestEval \cite{li2024latesteval} creates dynamic reading comprehension evaluations from recent texts using a three-step process: collecting texts, extracting key information, and constructing questions with template-filling or LLMs. WIKIMIA \cite{shi2023detecting} is a dynamic benchmark of post-2023 Wikipedia events, including paraphrased examples generated by ChatGPT. KIEval \cite{yu2024kieval} is an interactive framework with an LLM-powered "interactor" for multi-round dialogues to assess deep comprehension beyond mere recall. LiveCodeBench \cite{jain2024livecodebench} continuously collects new coding problems from LeetCode, AtCoder, and CodeForces for a contamination-free benchmark, revealing performance drops in some models, such as DeepSeek \cite{guo2024deepseek}. Termite \cite{ranaldi2024investigating} is a text-to-SQL dataset encrypted to prevent public access, designed to match the properties of the Spider dataset and address contamination observed in GPT-3.5. GSM1K \cite{zhang2024GSM1K} assesses the true reasoning ability of large language models by creating a new benchmark with similar style and complexity to GSM8k, revealing significant accuracy drops and evidence of memorization in many LLMs. LiveBench \cite{white2024livebench} introduces (1) frequently updated questions from recent sources, (2) automatic scoring based on ground-truth values, and (3) a variety of challenging tasks, including math, coding, reasoning, language, instruction following, and data analysis. It features questions from recent math competitions, arXiv papers, and news articles.
The frequently updated strategy ensures contamination-free results but leads to high evaluation costs.

Unlike the methods mentioned above, we categorize contamination into unintentional and malicious types.
We apply three decontamination rules to mitigate unintentional data leakage while collecting data from a broader domain.
Meanwhile, our MMLU-CF benchmark maintains the test set closed-source to prevent malicious data leakage.

\section{The MMLU-CF Benchmark}

\begin{figure*}[th]
\centering
\includegraphics[width=2\columnwidth]{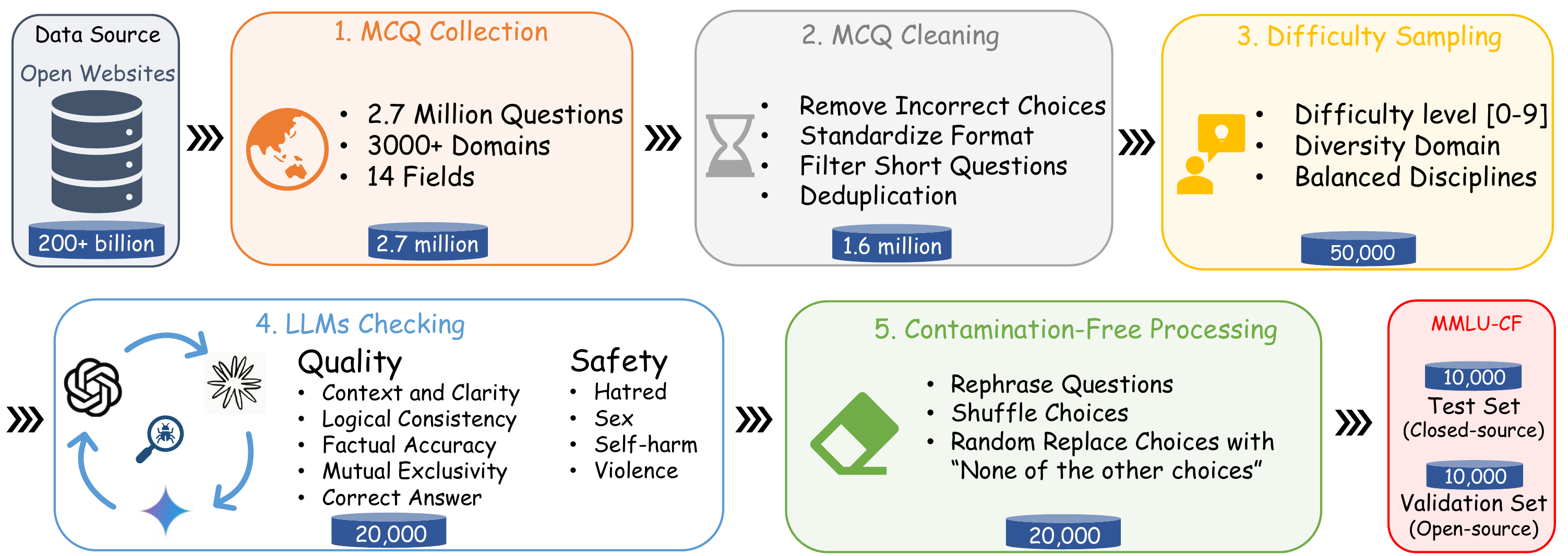}
\caption{The construction pipeline of the MMLU-CF Benchmark. The pipeline involves (1) MCQ Collection to gather a diverse set of questions; (2) MCQ Cleaning to ensure quality; (3) Difficulty Sampling to ensure an appropriate difficulty distribution for questions; (4) LLMs checking: The LLMs, including GPT-4o, Gemini, and Claude, are reviewing the accuracy and safety of the data; and (5) Contamination-Free Processing to prevent data leakage and maintain dataset purity. Ultimately, this process results in the MMLU-CF, consisting of 10,000 questions for the closed-source test set and 10,000 for the open-source validation set.}
\label{fig_3}
\end{figure*}

\subsection{Overview}
The MMLU-CF benchmark contains 20,000 data points and spans 14 fields, screened from 200+ billion documents on public open websites. To produce this diverse, high-quality, safety and contamination-free benchmark, we employ a series of steps, shown in Figure \ref{fig_3}. These steps include (1) MCQ Collection, (2) MCQ Cleaning, (3) Difficulty Sampling, (4) LLMs Checking, and (5) Contamination-Free Processing. Ultimately, we curate a dataset comprising 10,000 questions for the test set and 10,000 questions for the validation set respectively. The test set remains closed-source to prevent malicious exposure of the questions \cite{zhang2024GSM1K}, while the validation set is open-source to validate the authenticity and effectiveness of the questions. For more details on data statistics and prompt instructions, refer to the Appendix. The following sections outline the steps involved in processing the raw data.

\subsection{Dataset Construction Pipeline} 
\label{sec:dataset_construction_pipeline}
\textbf{(1) MCQ Collection.} Firstly, to preliminary mitigate the issue of our benchmark being exposed to the training data of large language models, we diversified the sources of our benchmark questions as much as possible. To achieve this, we leveraged over 200 billion documents from public open-source websites and employed rule-based methods to extract 2.7 million multiple-choice questions with answers as the raw questions. Unlike previous efforts, such as those by \cite{hendrycks2020mmlu, wang2024mmlupro}, which relied on a few sources to collect questions, these 2.7 million questions encompassed over 3,000 different website domains, ensuring a wide variety of content. These questions spanned 14 fields, including Health, Math, Physics, Business, Chemistry, Philosophy, Law, Engineering, and so on. 

\noindent\textbf{(2) MCQ Cleaning.} With the 2.7 million raw question points, we employed a series of filtering techniques for initial data cleaning. We first removed questions with choices number other than four. Next, we eliminated choices without content and analyzed the format of the choices. We then excluded questions with choices not labeled as A, B, C, or D, converted all choice labels to uppercase, and adjusted the answers accordingly. We filtered out questions with a length of less than 10 characters, used regular expressions to standardize answer formats, and removed original question numbering. After ensuring the correctness of choice order and question completeness, we conducted further checks on answer formats, removed redundant numbering, and ensured answers were within the provided choices. Finally, we eliminated Roman numeral labels, cleaned up question numbering, removed questions with lowercase initial letters and non-English characters, and performed deduplication. Through these steps, the data scale was reduced to 1.66 million.

\begin{figure}[h]
\centering
\includegraphics[width=0.48\textwidth]{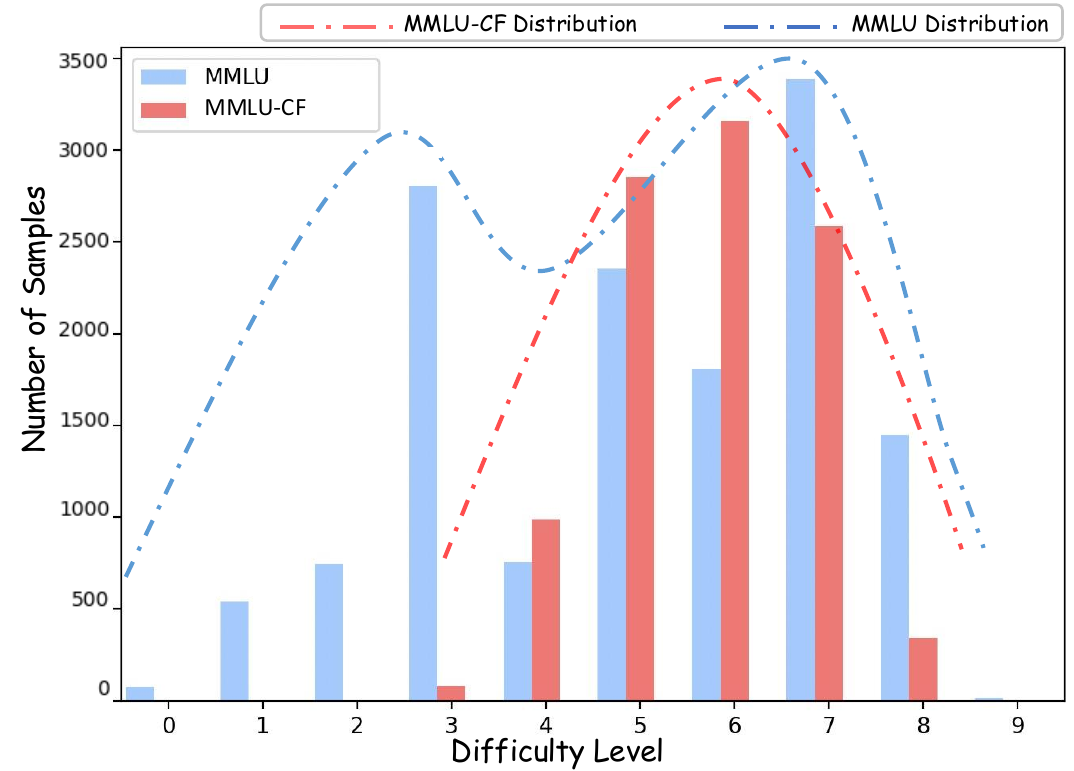}
\caption{The difficulty levels produced by GPT-4o for MMLU and MMLU-CF are analyzed. In our data, we randomly sampled 10,000 questions for visualization.}
\label{fig_4}
\end{figure}

\noindent\textbf{(3) Difficulty Sampling.} Due to the rapid advancement in the capabilities of LLMs, evaluations on MMLU \cite{hendrycks2020mmlu} have reached a bottleneck, indicating that the difficulty of the test set can no longer meet the needs of assessing the new generation of models. For instance, the latest frontier models, including GPT-4o, Gemini-1.5-Pro, and Claude, all published in early to mid-2024, have achieved accuracy rates ranging from 86\% to 88\%. Therefore, we aim to establish a more challenging benchmark to more effectively evaluate and drive progress in the new generation of models.

To investigate this further, we first used GPT-4o to categorize the difficulty levels of the original MMLU data. We employed the following query prompt for GPT-4o: ``Please rate the difficulty of this question on a scale of [0-9], where level [0] represents the easiest question and level [9] represents the most difficult.'' This resulted in a difficulty distribution for MMLU, as demonstrated in Figure \ref{fig_4}. Nearly one-third of the questions have a difficulty level below [4], and the abundance of easy questions is one of the reasons why LLMs achieve high scores on MMLU.

To categorize our data according to MMLU difficulty, we used the above difficulty levels of MMLU questions as a reference and then applied a 5-shot query for GPT-4o to classify the difficulty of 1.66 million clean questions. To ensure an appropriate level of difficulty, we selected questions using a normal distribution centered around a difficulty level of [6], as indicated in Figure \ref{fig_4}. During the sampling process, to ensure the diversity and quality of the questions, we maintained a balanced distribution of question categories, maximized the diversity of domains, and ensured that questions had corresponding explanations whenever possible. Finally, the 1.6 million questions reduce to 50,000.


\noindent\textbf{(4) LLMs Checking.} In the previous step, we selected 50,000 questions, ensuring moderate difficulty, domain diversity, and category balance, while also aiming to include explanations where possible. Although these questions were objectively accurate, having been sourced mostly from examination websites, further review for quality and harmlessness was necessary. Given the powerful capabilities of LLMs, these models are already employed in various fields such as data analysis \cite{zhao2024ltgc} and AI-driven decision-making \cite{zheng2024judging, chiang2024chatbot, yu2023wavecoder}. However, relying on a single model for review may introduce biases inherent to that LLM \cite{zheng2024judging}. To address the biases as much as possible, we employed three different LLMs, including GPT-4o, Gemini, and Claude, to review the quality and harmlessness of these MCQs.

For the quality of questions, we assessed them based on the following criteria:
\begin{itemize}
\vspace{-5px}
\item Context and Clarity: Are the question and choices consistent and unambiguous, providing enough context for understanding?
\vspace{-5px}
\item Logical Consistency: Are the question and choices logically structured without contradictions?
\vspace{-5px}
\item Factual Accuracy: Are the question and choices factually correct and not misleading?
\vspace{-5px}
\item Mutual Exclusivity: Are choices mutually exclusive without overlap?
\vspace{-5px}
\item Correct Answer: Is the correct answer included in the choices?
\end{itemize}

From the perspective of harmlessness, we reviewed the content from the following four aspects:
\begin{itemize}
\vspace{-5px}
\item Non-hatred: Ensure the content does not contain hate speech.
\vspace{-5px}
\item Non-sex: Ensure the content does not contain sexual suggestions or inappropriate sexual content.
\vspace{-5px}
\item Non-selfharm: Ensure the content neither contains self-harm nor encourages self-harm.
\vspace{-5px}
\item Non-violence: Ensure the content does not contain violence or incite violence.
\end{itemize}

Additionally, we used these three models to rate the questions on a scale from [1] to [5], where [5] represents the highest quality. Ultimately, we selected questions with an average score greater than [4] to construct test and validation sets of MMLU-CF. Then, inspired by Decontaminator \cite{yang2023rethinking}, we used GPT-4o to perform redundancy detection \cite{yang2023rethinking} on semantically identical test and validation questions. Furthermore, in our post-analysis, these questions came from over 1,000 web domains to ensure their diversity.

\begin{figure}[h]
\centering
\includegraphics[width=0.8\columnwidth]
{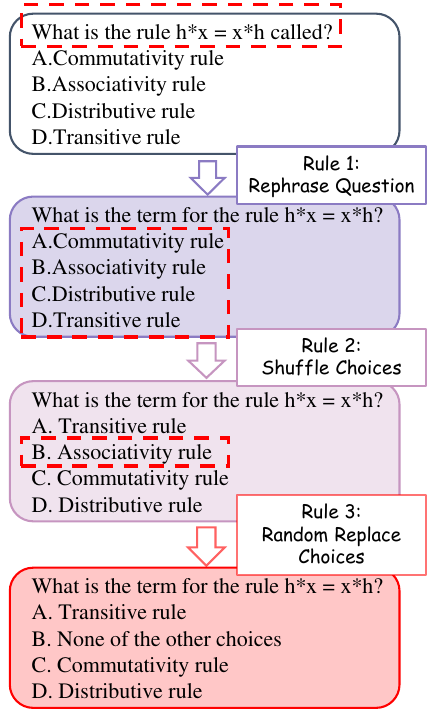}
\caption{A MCQ instance by Contamination-free Processing. The top box is the input MCQ, and the bottom box is the decontaminated MCQ.}
\label{fig_5}
\end{figure}

\noindent\textbf{(5) Contamination-Free Processing.} Moreover, to avoid unintentional contamination and to assess the LLMs' reasoning and understanding abilities rather than their memorization of answers \cite{carlini2023quantifying}, we implemented the following three decontamination rules as shown in Figure \ref{fig_5}:

(1) Rule 1: Rephrase Question. Rewriting questions helped reduce the model's dependence on previously encountered training data \cite{zhu2024inference}, thereby mitigating performance inflation caused by the models memorizing leaked benchmarks.

(2) Rule 2: Shuffle Choices. To prevent the model from answering correctly based on memorizing the sequence of choices, we shuffled the choices \cite{gupta2024changing}. If the last option was `None of the above' or `All of the above,' we only shuffled the first three choices.

(3) Rule 3: Random Replace Choices. We randomly replaced one of the choices in the question with `None of the other choices' with a 50\% probability. If the last option was `None of the above' or `All of the above', we skipped this question. When replacing the correct option, it remained a valid choice, requiring the model to use more reasoning to answer the question. Similarly, when replacing an incorrect option, it acted as a distractor, necessitating more comprehension and reasoning from the model to answer correctly.

These rules help mitigate both malicious and unintentional leakage to varying degrees. After that, we divided the data into 10,000 validation and 10,000 test sets, maintaining similar difficulty and categories across both sets. The test set was kept closed-source to prevent malicious contamination. 


\section{Experiments}

\begin{table*}[!h]
\scriptsize
\centering
\begin{tabular}{p{0.6cm}l|l|cccccc}
\toprule
& \multirow{3}{*}{Model} & \multirow{1}{*}{MMLU} & \multicolumn{3}{c}{MMLU-CF } & \multicolumn{3}{c}{MMLU-CF} \\
& & 5-shot (\%) & \multicolumn{3}{c}{5-shot (\%)} & \multicolumn{3}{c}{0-shot (\%)} \\
\cmidrule(r){4-6} \cmidrule(r){7-9}
    & & Test & Test & Validation & $\Delta$ (\%) & Test & Validation & $\Delta$ (\%) \\
    \midrule
    \multirow{7}{*}{API} 
    & GPT-4o \cite{openai2024gpt4o} & 88.0 & 73.4 & 73.4 & +0.0 & 71.9 & 72.4 &-0.5 \\
    & GPT-4-Turbo \cite{achiam2023gpt}&86.5 & 70.4 & 70.1 & +0.3 &68.9 & 68.7 &+0.1\\
    & GPT-4o-mini \cite{gpt4omini} &81.8 & 65.5 & 65.1 & +0.4 &66.0 & 65.3 &+0.7\\
   & Gemini-1.5-Flash \cite{reid2024gemini}&78.7  & 64.8 & 64.9 & -0.1 & 56.7 & 56.9 &-0.2 \\
    & GPT-3.5-Turbo \cite{gpt35}&71.4 & 58.2 & 59.0 & -0.8 & 57.2 & 58.1 &-0.9 \\
 
    \midrule
    \multirow{6}{*}{Large} 
    & Qwen2.5-72B-instruct \cite{qwen2.5}&85.3 &71.6 &71.3 &+0.3 & 70.6 &70.4 &+0.2 \\
    & Llama-3-70B-instruct \cite{meta2024llama3}&82.0 & 68.9 & 68.8 & +0.1 & 68.1 & 67.4 &+0.7 \\
    & Llama-3.3-70B-instruct \cite{meta2024llama3}&86.3 & 68.8 & 67.8 & +1.0 & 67.6 & 67.5 &+0.1 \\
    & Llama-3.1-70B-instruct \cite{meta2024llama3}&$86.0^\ddag$  &  68.7 & 68.1 & +0.6 & 70.4 & 69.7 &+0.7 \\
    & Phi-3.5-MoE-instruct  \cite{abdin2024phi}&78.9 & 64.6 & 64.5 & +0.1 & 63.1 &  62.1 &+1.0 \\
    & Qwen2-72B-instruct \cite{bai2023qwen} &82.3 & 63.7 & 64.3 & -0.6 & 62.4 & 62.5 &-0.1 \\
    & Mixtral-8x22B-instruct \cite{jiang2024mixtral}&76.2 & 62.8 & 62.5 & +0.3 & 65.3 & 64.8 &+0.5 \\
    & Qwen1.5-72B-chat \cite{bai2023qwen} &75.6 & 59.8 & 60.2 & -0.4 & 59.1 & 59.6 &-0.5 \\
    & Llama-2-70B-chat \cite{meta2024llama3}&68.9 & 52.2 &  51.8 & +0.4 & 51.2 & 50.9 &+0.3 \\
    \midrule
    
    \multirow{5}{*}{Medium} 
    & Qwen2.5-32B-instruct \cite{qwen2.5}&$83.9^\dag$ &69.7 &68.8 &+0.9 & 68.9 &68.8 &+0.1 \\
    & Phi-4-14B \cite{abdin2024phi4technicalreport}&84.8 &67.8  &68.5 &-0.7 &68.5  &69.4 &-0.9 \\
    & Qwen2.5-14B-instruct \cite{qwen2.5}&79.9 &66.4 &66.1 &+0.3 &67.0 & 66.0 &+1.0 \\
    & Phi-3-medium-instruct \cite{abdin2024phi}& 77.9 & 64.2 & 64.2 & +0.0 & 62.5 & 62.7 &-0.2 \\
    & Gemma2-27B\cite{team2024gemma} &75.2  &63.9 &63.5  &+0.4 & 64.2 &64.0 &+0.2 \\
    & Yi-1.5-34B-chat \cite{young2024yi} &76.8 & 61.3 & 60.5 &+0.8 & 60.6 &  59.5 &+1.1\\
    & Mixtral-8x7B-instruct-v0.1 \cite{jiang2024mixtral} &70.5 & 58.3 & 57.1 & -1.2 & 58.9 &  58.5 &+0.4 \\
    & Deepseek-v2-lite-chat \cite{deepseekv2}&55.7 & 49.3 & 48.7 & +0.6 & 48.2 & 47.7 &+0.5 \\
    & Baichuan-2-13B-chat \cite{yang2023baichuan}&57.3 & 48.3 & 48.6 & -0.3 & 47.1 &  48.1 &-1.0 \\
    & Llama-2-13B-chat \cite{touvron2023llama}&54.8 &  42.8 & 42.1 & +0.7 &  44.8 & 44.6 &+0.2 \\

\midrule
\multirow{12}{*}{Small} 
& Qwen2.5-7B-instruct \cite{qwen2.5} &	$75.4^\dag$ &61.3 &60.4 &+0.9 &59.3 & 58.6 &+0.7 \\
& Qwen2-7B-instruct \cite{bai2023qwen} &70.5 & 58.1 & 57.9 & +0.2 & 58.3 & 57.4 &+0.9 \\
& Glm-4-9B-chat \cite{glm2024chatglm} & 72.4 & 57.8 &57.9 &-0.1 &58.6 &58.7 &-0.1 \\
& Internlm-2.5-7B-chat \cite{cai2024internlm2} &72.8 & 57.3 & 56.8 & +0.5 & 57.9 & 56.9 &+1.0 \\
& Llama-3-8B-instruct \cite{meta2024llama3}&68.4 &57.3 & 56.5 & +0.8 & 56.4 & 55.4 &+1.0 \\
& Llama-3.1-8B-instruct  \cite{meta2024llama3}&68.1& 57.1 & 57.9 & -0.8 &56.1 & 56.1 &+0.0 \\
& Gemma-2-9B \cite{team2024gemma} &71.3 &53.7 &53.3 & +0.4 & 32.1 & 31.2 &+0.9 \\
& Yi-1.5-6B-chat \cite{young2024yi} &62.8 &52.8 &51.4 & +1.4 & 52.2 & 51.9 &+0.3 \\
& Mistral-7B-instruct-v0.3 
\cite{jiang2023mistral}&60.3  & 50.7 & 50.9 & -0.2 & 51.1 & 50.9 &+0.2 \\

& Baichuan-2-7B-chat \cite{yang2023baichuan} &52.9 & 44.5 & 43.9 & +0.6 & 43.9 & 44.0 &-0.1 \\
& Llama-2-7B-chat \cite{touvron2023llama}&45.3 & 39.4 & 38.5 & +0.9 & 41.9 & 40.9 &+1.0\\
\midrule
\multirow{4}{*}{Mini} 
& Phi-3-mini-instruct (3.8B) \cite{abdin2024phi}&70.9 &57.9 & 58.1 & -0.2 & 58.2 & 57.5 &+0.7 \\
& Phi-3.5-mini-instruct (3.8B) \cite{abdin2024phi}&69.1 & 57.9 & 57.4 & +0.5 & 58.3 & 57.7 &+0.6 \\
& Qwen2.5-3B-instruct \cite{qwen2.5}&$64.4^\dag$&55.9 &56.4 &-0.5 &54.3 &53.9 &+0.4 \\
& Qwen2.5-1.5B-instruct \cite{qwen2.5}&$50.7^\dag$ &51.2 &51.0 & +0.2 &50.7 & 50.4 &+0.3 \\
& Qwen2-1.5B-instruct \cite{bai2023qwen}&52.4& 47.1 & 47.5 & -0.4 & 45.2 & 44.5 &+0.7 \\
& Gemma-2-2B \cite{team2024gemma}&51.3 & 43.9 & 42.4 & +1.5 & 30.5 & 29.4 & +0.9 \\
& Qwen2.5-0.5B-instruct \cite{qwen2.5} &$24.1^\dag$ &41.9 &41.1 &+0.8 &36.0  & 34.9 &+1.1 \\
& Internlm-2-chat-1.8b \cite{cai2024internlm2}&47.1  & 40.5 & 39.4 & +1.1 & 41.2 & 39.8 &+1.4 \\
& Qwen2-0.5B-instruct \cite{bai2023qwen} &37.9  & 38.3 & 38.3 & +0.0 & 33.5 & 33.5 &+0.0 \\
\bottomrule
\end{tabular}
\caption{
Performance of various models on MMLU and MMLU-CF (ours).
Both 0-shot and 5-shot evaluations don't employ COT~\cite{kojima2022large}, except for additional explanations.
$\Delta$ means the absolute score difference of models between validation and test sets.
$\ddag$ denotes 0-shot with COT.
$\dag$ indicates employing MMLU-redux~\cite{gema2024mmluredux}, the results are from Qwen2.5 homepage \cite{qwen2.5}.
}
\label{tab:results}
\vspace{-10px}
\end{table*}

\subsection{Evaluation Models}
We evaluate 40+ models across various sizes by the evaluation platform OpenCompass \cite{2023opencompass}, including open-source models ranging from 0.5B to 72B and closed-source APIs. The experiments include models with different classes, such as GPTs \cite{achiam2023gpt} (GPT-4o (v2024-10-1), GPT-4o-mini (v2024-10-1), GPT-4-Turbo (v2024-2-15), GPT-3.5-Turbo (v2024-2-15)), Gemini \cite{reid2024gemini} (Gemini-1.5-Flash), and public models like Llama-3-\{8, 70\}B-chat \cite{meta2024llama3}, Llama-3.1-\{8, 70\}B-chat \cite{meta2024llama3}, Mixtral-\{7, 8x7, 8x22\}B-instruct,  Phi-4 \cite{abdin2024phi4technicalreport}, Phi-3.5-\{mini, small\} \cite{abdin2024phi}, Gemma-2-\{2, 9, 27\}B \cite{team2024gemma}, Qwen2.5-\{0.5, 1.5, 7, 14, 70\}B \cite{qwen2.5}.

\subsection{Evaluation Metrics}
We employ both 5-shot and 0-shot approach to measure the performance of large language models on the MMLU-CF test and validation set. Additionally, we categorize the open-source models based on their parameter size into four sections: Large ($>$50B), Medium (13B-50B), Small (6B-12B), and Mini (0.5-5B). The $\Delta$ is the absolute score difference between the test and validation sets.

\subsection{Evaluation Methods for Public}
Two evaluation methods are supported for our benchmark. The users could voluntarily submit evaluation requests by providing Hugging Face open-source model types or API formats through the introduction of our project homepage. Besides, we will actively evaluate the latest popular models from Hugging Face as well as mainstream APIs.

\subsection{Results and Analysis}

As shown in Table \ref{tab:results}, GPT-4o emerges as the strongest model across both close-sourced and open-sourced models, achieving a score of 73.4\% in the 5-shot test and 71.9\% in the 0-shot test on test set. This result highlights GPT-4o’s ability to handle a wide range of tasks effectively and serves as the benchmark for other models.

Among the API-based models, GPT-4-Turbo achieves 70.4\% in the 5-shot test and maintains a robust performance of 68.9\% in the 0-shot test. Notably, Gemini-1.5-Flash delivers competitive performance at 64.8\% in the 5-shot test but lags behind GPT-4 variants.

In the large-model category, Qwen2.5-72B-instruct outperforms its peers with a strong 71.6\% in the 5-shot test and a slight improvement of +0.3\% between test and validation scores. Llama-3.3-70B-instruct also delivers consistent performance, though slightly behind Qwen2.5. 

Within medium models, Qwen2.5-32B-instruct stands out with 69.7\% in the 5-shot test, significantly outperforming other models in this category. Meanwhile, Phi-4-14B continues to excel with a strong 67.8\% in 5-shot and 68.5\% in 0-shot, maintaining its dominance even over some larger models, reflecting its efficiency and robustness.

In the small-model category, Qwen2.5-7B-instruct performs well, achieving 61.3\% in the 5-shot test. Both outperform many other small and even medium-sized models. 

Among mini-sized models, Phi-3.5-mini-instruct 
with 3.8B achieves the best performance with 57.9\% in the 5-shot test. Qwen2.5-3B-instruct closely follows with 55.9\%. 



\begin{table}[htbp]
\small
\centering
\setlength{\tabcolsep}{2pt}
\begin{tabular}
{p{0.65cm}p{0.65cm}p{0.65cm}ccc}
\hline
{Rule 1} & {Rule 2} & {Rule 3} & {GPT-4o} & {GPT-3.5-Turbo} & {Llama-3.1-8b} \\
\hline
- & - & - & 79.8 &  65.3 &  63.8 \\
\checkmark & - & - & 78.6 & 63.1 & 62.3 \\
\checkmark & \checkmark & - & 77.9 & 62.8 & 61.8 \\
\checkmark & \checkmark & \checkmark & 73.4 & 58.2 & 57.1 \\
\hline
\end{tabular}
\caption{5-shot results of applying different decontamination rules to MMLU-CF test set.}
\label{tab:decon}
\vspace{-10px}
\end{table}

\subsection{Properties of Partitioning Test and Validation Sets}
We partition the benchmark dataset into test and validation sets, then calculate the absolute score difference as $\Delta$ for LLMs, it not only helps prevent test set leakage but also offers the following benefits: Firstly, as shown in Table \ref{tab:results}, before the validation set is publicly released, about 60\% of \(\Delta\) values are less than 0.5, and 96\% of \(\Delta\) values are below 1.0. This indicates that the evaluation results of LLMs are significantly consistent across the test and validation sets, demonstrating the effectiveness of the validation set in evaluating model generalization. Once the validation set is made public, potential data leakage can cause the models to overfit on the validation set, leading to an increase in \(\Delta\) values. Thus, the design of \(\Delta\) serves as a method to monitor whether benchmarks might be compromised. This approach helps ensure the fairness and integrity of the benchmarks, preventing models from exploiting leaked data to artificially enhance their performance.

\subsection{Ablation Study on Different Decontamination Rules}
We conducted ablation experiments on the MMLU-CF to evaluate the performance of different models under three modification rules: Rephrase Question (Rule 1), Shuffle Choices (Rule 2), and Random Replace Choices (Rule 3) with ``None of the other choices''. The LLMs used in this study are GPT-4o, GPT-3.5-Turbo, and Llama-3-8b. The experimental results are summarized in Table \ref{tab:decon}, the rule 1 causes a slight decrease in performance across all models. However, the addition of rule 2 and rule 3 results in a more significant decline, particularly when all three rules are applied. This suggests that the later rules either remove more valuable data or create a cumulative effect that further hampers model performance. 
The significant performance drop on MMLU-CF demonstrates the effectiveness of the three decontamination rules, particularly rule 1 and rule 2, which don't alter the difficulty of the original questions.
Additionally, the more pronounced drop observed in GPT-3.5-Turbo and Llama-3-8b suggests that smaller models are more sensitive to the removal of potentially useful data or the added complexity from these rules, making them less effective under stricter decontamination.


\section{Conclusion}
In this paper, we propose and construct MMLU-CF, a contamination-free and challenging multiple-choice question benchmark, to reassess large language models' understanding of world knowledge. Specifically, we categorize contamination into unintentional and malicious types. To prevent unintentional data contamination, we design three decontamination rules to mitigate unintentional data leakage while collecting data from a broader domain. To prevent malicious data contamination, we keep the test set closed-source while making the validation set publicly available for transparency. Evaluation results demonstrate that GPT-4o achieved a 5-shot score of 73.4\%, ranking at the top among evaluated models. This result is significantly lower than the 88.0\% on MMLU, highlighting the challenging and contamination-free nature of MMLU-CF. We believe this benchmark would promote fair model evaluation and provide valuable insights for the design of future contamination-free benchmarks.

\section{Limitations}
Although this dataset is constructed with the utmost objectivity and fairness, leveraging multiple large language models to verify the correctness of the questions and answers, it is still possible that some errors may remain. To address this, we have provided a validation set that is available to the public for further scrutiny and verification. Additionally, this dataset primarily focuses on multiple-choice questions and language modalities. However, other aspects of large models' capabilities, such as math and code reasoning, multi-modal understanding (e.g., image and audio), and specific domain expertise, still require evaluation with similarly unbiased and contamination-free benchmarks. 

\section{Ethics Statement}
MMLU-CF was created using open-source data and methodologies to ensure transparency. The benchmark is designed to provide fair and reliable evaluations through decontamination rules and verification. While efforts were made to minimize errors, some may remain, and we encourage the community to review the publicly available validation set for further improvements. This benchmark focuses on language modalities, and future work is needed for unbiased evaluation in other areas. We call for the responsible use of this dataset to promote ethical and equitable AI development.

\bibliography{acl_latex}
\clearpage
\appendix

\clearpage

\section{Appendix}
\label{sec:appendix}

\subsection{Disciplinary Distribution of MMLU-CF}

\begin{figure}[ht]
\small
\centering
\includegraphics[width=0.9\columnwidth]{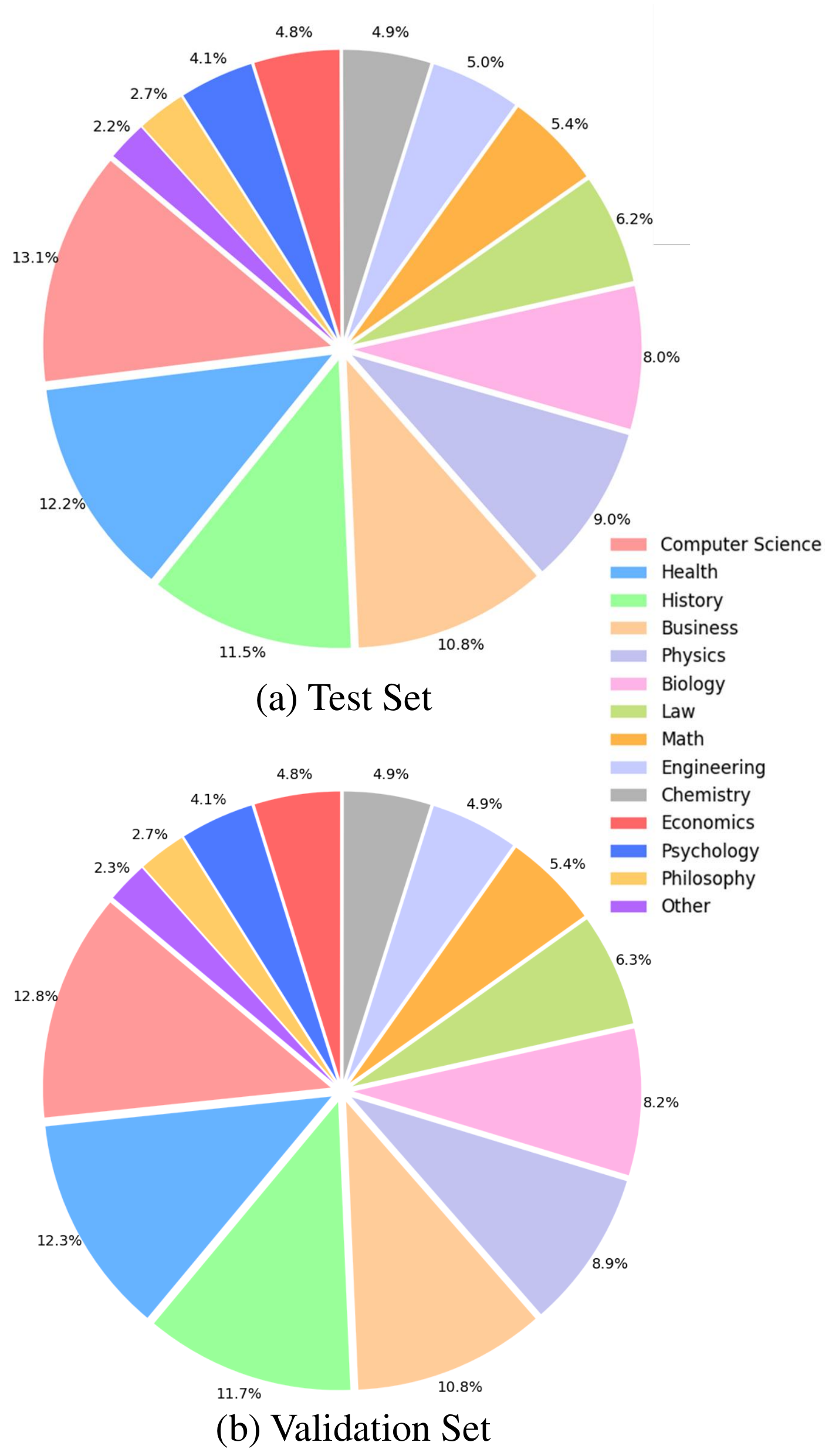}
\caption{Distribution of Disciplines in MMLU-CF.}
\label{Sup_1}
\end{figure}

The Figure \ref{Sup_1} demonstrates the visualization of MMLU-CF test and validation sets. We find that their disciplinary distribution proportions are quite similar. The most prevalent disciplines are Computer Science, Health, and History, with proportions of 13.1\%, 12.2\%, and 11.5\% in the test set, respectively. This distribution may lead to slight differences in the performance of different models. Table \ref{Sup_1}, we present the performance of various large models across different disciplines. GPT-4o achieves the best performance in terms of average accuracy and different disciplines. We observe that the models perform worst in Computer Science. This is because the domain not only requires fundamental knowledge of Computer Science but also involves code understanding, which increases the difficulty. Qwen2.5-72B, -32B brings new upgrades in mathematics and coding, delivering the best results in mathematics, engineering, and computer science. Despite its small size, Phi-4 achieves competitive results compared to larger models, showcasing its efficiency in handling complex tasks.

\subsection{The Effect of Decontamination Rules} 
\begin{figure}[htbp]
\centering
\includegraphics[width=0.8\columnwidth]{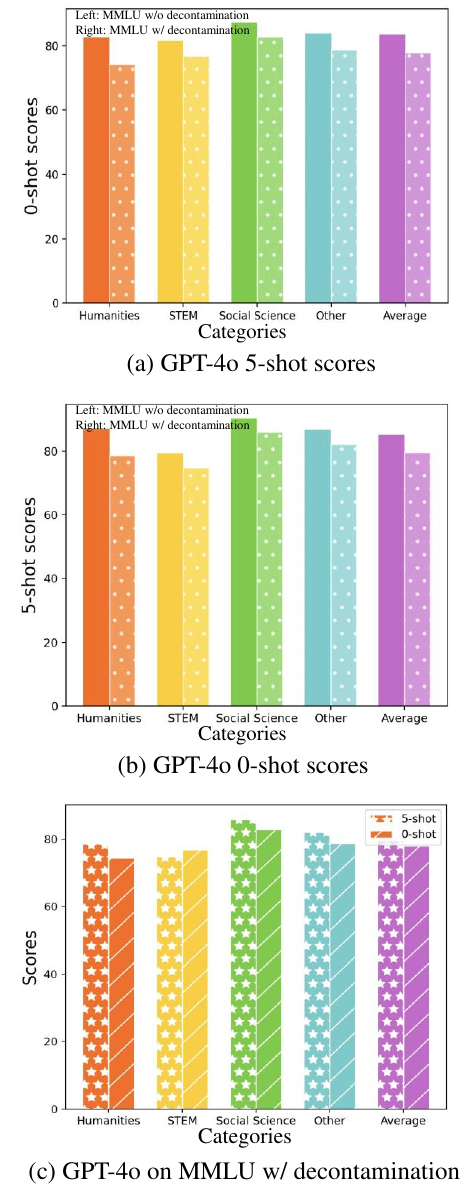}
\caption{GPT-4o evaluation comparison on MMLU with and without our decontamination rules.}
\label{fig_6}
\vspace{-10px}
\end{figure}

In the methods section, we presented three types of question modification rules applied to the MMLU-CF dataset: question rephrasing, shuffling choices, and randomly replacing an option with ``None of the other choices.'' To validate the effectiveness of these modifications, we first applied these three rules to the MMLU \cite{hendrycks2020mmlu}. The results, shown in Figure \ref{fig_6}, indicate that these modifications lead to a decrease in 5-shot and 0-shot scores for GPT-4o. Furthermore, when comparing these results to those on the MMLU-CF dataset, as depicted in Table \ref{tab:decon}, the accuracy drop is more pronounced on the MMLU dataset. This suggests a higher likelihood of data leakage in large models when using the MMLU dataset. In contrast, the MMLU-CF dataset, due to its broad and closed-source nature, exhibits a lower risk of data leakage.

\begin{table*}[htbp]
\centering 
\scriptsize
\begin{tabular}{lcccccc} 
\toprule
Subject         & GPT-4o    & GPT-4o-mini   & Llama-3.3-70B & Qwen2.5-72B& Qwen2.5-32B &Phi-4-14B\\
~ & \cite{openai2024gpt4o} & \cite{gpt4omini} & \cite{meta2024llama3} &  \cite{qwen2.5} &\cite{qwen2.5} &\cite{abdin2024phi4technicalreport}
\\ \midrule
Math            &  56.09   & 45.83   &56.3 &\textbf{67.51} & 63.10  & 62.18    \\
Physics         &  \textbf{75.15}   & 64.47  &69.0 &74.00 & 71.12  &69.15     \\
Chemistry       &  \textbf{72.44}   & 66.54  &68.3 &69.62 & 68.81  &67.13     \\
Law             &  \textbf{81.46}   & 72.73  &73.6 &75.15 & 72.55  &71.84     \\
Engineering     &  60.15   & 55.41  &56.7 &\textbf{61.39} & 57.69  &54.67     \\
Economics       &  \textbf{78.33}   & 66.31  &72.5 &74.95 & 68.90  &68.88     \\
Health          &  \textbf{81.09}   & 76.11  &79.3 &80.23 & 78.55  &76.29     \\
Psychology      &  \textbf{80.10}   & 70.28  &77.5 &78.95 & 77.94  &75.45    \\
Business        &  70.90   & 63.81  &64.7 &\textbf{71.00} & 68.69  &65.19     \\
Biology         &  \textbf{82.84}   & 74.63  &75.5 &78.88 & 74.53  &75.91     \\
Philosophy      &  \textbf{81.82}   & 77.99  &78.9 &74.24 & 72.73  &76.08     \\
Computer Science&  55.50   & 51.09  &51.0 &56.12 & \textbf{68.79}  &51.09     \\
History         &  \textbf{77.23}   & 67.05  &71.2 &71.19 & 68.79  &68.09     \\
Other           &  \textbf{74.83}   & 64.74  &67.9 &68.15 & 66.88  &65.55     \\
\midrule
Average        & \textbf{73.42}  &65.52 & 68.82 & 71.60& 68.81  &67.68     \\
\bottomrule
\end{tabular}
\caption{Performance of different models on MMLU-CF discipline under a 5-shot test set. The best result is emphasized in bold.}
\label{tab:mmlu_performance}
\vspace{-10px}
\end{table*}

\subsection{Prompt Used for LLMs Checking}
Table \ref{table_prompt} shows the prompt used in the LLMs checking processing to verify the correctness of questions. For safety, we used GPT-4’s built-in safety filter under the strongest constrains to filter out unsafe content related to hate speech, sexual content, self-harm, and violence.

\begin{table}[h!]
\centering
\small
\begin{tabular}{p{7.4cm}}
    \toprule
    $[$Instruction$]$\\
    Please review the following question and corresponding choices for correctness based on these criteria: \\ 
    \textbf{Context and Clarity}: Are the question and choices consistent and unambiguous, providing enough context for understanding? \\ 
    \textbf{Logical Consistency}: Are the question and choices logically structured without contradictions? \\ 
    \textbf{Factual Accuracy}: Are the question and choices factually correct and not misleading? \\ 
    \textbf{Mutual Exclusivity}: Are choices mutually exclusive without overlap? \\ 
    \textbf{Correct Answer}: Is the correct answer included in the choices?\\
    $[$Question to be reviewed$]$\\
    \{question\} \\ 
    $[$Choice to be reviewed$]$\\
    \{choice\} \\ 
    $[$Response$]$\\
    Rate the question's correctness on a scale of 1 to 5, with 5 being correct; Only give an overall Rating. For example, Rating: 5 \\ 
    \bottomrule
\end{tabular}
\caption{The Prompt of LLMs Checking.}
\label{table_prompt}
\end{table}

\subsection{The Difficulty Level of MMLU-CF}

\begin{figure}[h]
\centering
\includegraphics[width=0.9\columnwidth]{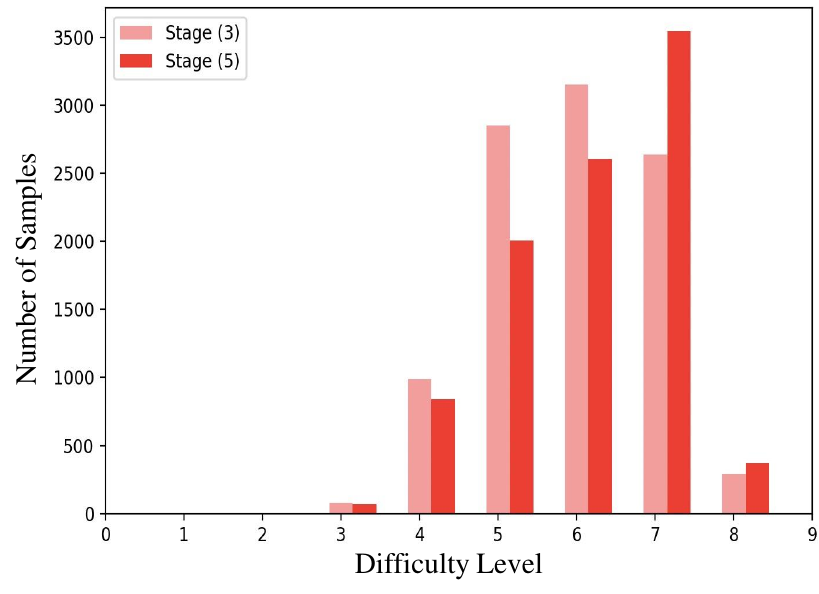}
\caption{The difficulty level distribution of MMLU-CF after stage (3) and (5).}
\label{Sup_2}
\end{figure}

Figure \ref{Sup_2} demonstrates the difficulty distribution of samples in the MMLU-CF dataset at various stages. In step three, we sampled normally around a difficulty level of 6. After applying the decontamination process in step five, we observed a notable change: the proportion of samples with difficulty level 5 significantly decreased, while the number of questions with difficulty level 7 increased. This indicates that the decontamination process introduced more challenging questions into the dataset, which meets the expectation.


\subsection{Sampled Questions for Different Disciplines}
In Table \ref{table_question1}, \ref{table_question2}, \ref{table_question3}, \ref{table_question4} and \ref{table_question5}, we present the questions from the validation set across various disciplines. For each subject, we have randomly sampled three questions for demonstration, which offers insights into the diversity and characteristics of the questions used in the validation process.
For more questions, we will publicly the validation set soon. 

\begin{table*}[h!]
\small
\centering
\begin{tabular}{|p{14cm}|}
\hline
\multicolumn{1}{|c|}{\textbf{Biology}} \\ \hline

\textbf{Question 1} \\
Which group of biological molecules is the most diverse in function? \\
A. Carbohydrates \quad B. Proteins \quad \\
C. Nucleic acids \quad D. Lipids \\
\textbf{Answer:} B \\ 
\midrule

\textbf{Question 2} \\
Which of these structures is the smallest? \\
A. Hydrogen atom \quad B. None of the other choices \quad \\
C. Mitochondrion \quad D. Viriod \\
\textbf{Answer:} A \\ 
\midrule

\textbf{Question 3} \\
Which of the following controls and regulates life processes? \\
A. Reproductive and endocrine systems \quad B. Endocrine and digestive systems \\
C. None of the other choices \quad D. Nervous and endocrine systems \\
\textbf{Answer:} D \\ 
\hline

\multicolumn{1}{|c|}{\textbf{Chemistry}} \\ \hline

\textbf{Question 1} \\
What occurs when silver chloride is exposed to sunlight? \\
A. Silver metal and chlorine gas are formed \quad B. Silver metal and hydrogen gas are formed \\
C. Only hydrogen gas is formed \quad D. Only silver metal is formed \\
\textbf{Answer:} A \\ 
\midrule

\textbf{Question 2} \\
What is the phenomenon called when a beam of light passes through a colloidal solution? \\
A. Cataphoresis \quad B. Tyndall effect \quad \\
C. Electrophoresis \quad D. Coagulation \\
\textbf{Answer:} B \\ 
\midrule

\textbf{Question 3} \\
Electrolytes play a crucial role in the chemistry of living organisms. What defines an electrolyte? \\
A. Contains electrodes \quad B. Conducts electricity when melted or put into solution \\
C. Generates light when electricity is applied \quad D. Contains electrons \\
\textbf{Answer:} B \\ 
\hline

\multicolumn{1}{|c|}{\textbf{Computer Science}} \\ \hline

\textbf{Question 1} \\
Which of the following is not a valid floating point literal in Java? \\

A. 5.0e2 \quad B. 033D \quad \\
C. 6.8 \quad D. 4.5f \\
\textbf{Answer:} B \\ 
\midrule

\textbf{Question 2} \\
\begin{verbatim}
#include <stdio.h>
int main() {
    int a = -1, b = 4, c = 1, d;
    d = ++a && ++b || ++c;
    printf("%d, %d, %d, %d\n", a, b, c, d);
    return 0;
}
\end{verbatim}
A. 0, 5, 2, 1 \quad B. 0, 4, 2, 1 \\
C. None of the other choices \quad D. 1, 4, 1, 1 \\
\textbf{Answer:} B \\ 
\midrule

\textbf{Question 3} \\
In what aspect did a digital computer not surpass an analog computer? \\
A. Accuracy \quad B. Reliability \quad \\
C. Speed \quad D. None of the other choices \\
\textbf{Answer:} A \\ 
\hline
\end{tabular}
\caption{Three Random Questions from the Biology, Chemistry and Computer Science of the MMLU-CF Validation Set.}
\label{table_question1}
\end{table*}

\begin{table*}[h!]
\small
\centering
\begin{tabular}{|p{14cm}|}
\hline
\multicolumn{1}{|c|}{\textbf{Engineering}} \\ \hline

\textbf{Question 1} \\
What functions can a diode perform? \\
A. Rectifier \quad B. None of the other choices \quad \\
C. Demodulator \quad D. Modulator \\
\textbf{Answer:} C \\ 
\midrule

\textbf{Question 2} \\
What is a periodic signal? \\
A. May be represented by g(t) = g(t + T0) \quad B. Value may be determined at any point \\
C. Repeats itself at regular intervals \quad D. All of the above \\
\textbf{Answer:} D \\ 
\midrule

\textbf{Question 3} \\
What are the advantages of using electron beam welding? \\
A. Absence of porosity \quad B. Welds are clean \quad \\
C. Distortion less \quad D. All of these \\
\textbf{Answer:} B \\ 
\hline

\multicolumn{1}{|c|}{\textbf{Math}} \\ \hline

\textbf{Question 1} \\
What is the result when \(\frac{1}{\sqrt{7}-2}\) is rationalized? \\
A. \((\sqrt{7}-2)/3\) \quad B. \((\sqrt{7}+2)/45\) \quad \\
C. \((\sqrt{7}+2)/5\) \quad D. \((\sqrt{7}+2)/3\) \\
\textbf{Answer:} D \\ 
\midrule

\textbf{Question 2} \\
What is the percentage increase in the area of a rectangle if each side is increased by 20\%? \\
A. 46\% \quad B. 44\% \quad \\
C. 42\% \quad D. 40\% \\
\textbf{Answer:} B \\ 
\midrule

\textbf{Question 3} \\
What is the radius of a sphere with a surface area of 616 cm²? \\
A. 21 cm \quad B. 7 cm \quad \\
C. 3.5 cm \quad D. 14 cm \\
\textbf{Answer:} B \\ 
\hline

\multicolumn{1}{|c|}{\textbf{Physics}} \\ \hline

\textbf{Question 1} \\
Daylight color film is calibrated for what type of light? \\
A. 3200 K \quad B. 3400 K \quad \\
C. 3000 K \quad D. 5400 K \\
\textbf{Answer:} D \\ 
\midrule

\textbf{Question 2} \\
On a Force versus position (F vs. x) graph, what signifies the work done by the force F? \\
A. The product of the maximum force times the maximum x \quad B. The length of the curve \\
C. The slope of the curve \quad D. The area under the curve \\
\textbf{Answer:} D \\ 
\midrule

\textbf{Question 3} \\
What is the phase difference between the voltage and current in a capacitor in an AC circuit? \\
A. $\pi$/3  \quad B. $\pi$/2 \quad \\
C. $\pi$  \quad D. 0 \\
\textbf{Answer:} B \\ 
\hline
\end{tabular}
\caption{Three Random Questions from the Enigineering, Math and Physics of the MMLU-CF Validation Set.}
\label{table_question2}
\end{table*}

\begin{table*}[h!]
\small
\centering
\begin{tabular}{|p{14cm}|}
\hline
\multicolumn{1}{|c|}{\textbf{Business}} \\ \hline

\textbf{Question 1} \\
Beth is the project manager for her organization. While her current project has numerous deliverables identified broadly, the specific details of these deliverables remain unclear. Beth is meticulously planning only the activities that are immediately forthcoming in the project. What is this project management planning approach called? \\
A. Rolling wave planning \quad B. Imminent activity management \quad \\
C. None of the other choices \quad D. Predecessor-only diagramming \\
\textbf{Answer:} A \\ 
\midrule

\textbf{Question 2} \\
How do you format Pivot Table report summary data as currency? \\
A. Type in the currency symbol \quad B. Use custom calculation \quad \\
C. Modify the field settings \quad D. None of the above \\
\textbf{Answer:} C \\ 
\midrule

\textbf{Question 3} \\
Which one of these choices is not considered an operating cost? \\
A. Maintenance cost \quad B. Salaries of high officials \quad \\
C. None of the other choices \quad D. Salaries of operating staff \\
\textbf{Answer:} B \\ 
\hline

\multicolumn{1}{|c|}{\textbf{Economics}} \\ \hline

\textbf{Question 1} \\
Which tax proposal did the Finance Minister announce the withdrawal of on 8th March following nationwide protests? \\
A. Tax on High Income Farmers \quad B. Tax proposal on EPF \quad \\
C. Kisan Kalyan Cess \quad D. All of above \\
\textbf{Answer:} B \\ 
\midrule

\textbf{Question 2} \\
In economics, what does the demand for a good indicate regarding the quantity that people: \\
A. None of the other choices \quad B. Need to achieve a minimum standard of living \\
C. Will buy at alternative income levels \quad D. Would like to have if the good were free \\
\textbf{Answer:} A \\ 
\midrule

\textbf{Question 3} \\
What is it called when a firm's supply rises as a result of implementing advanced technology? \\
A. Expansion in supply \quad B. Increase in quantity supplied \\
C. Contraction in supply \quad D. Increase in supply \\
\textbf{Answer:} D \\ 
\hline

\multicolumn{1}{|c|}{\textbf{Health}} \\ \hline

\textbf{Question 1} \\
Thrombocytes are more accurately referred to as \_\_\_\_\_\_? \\
A. Megakaryoblasts \quad B. Clotting factors \quad \\
C. Megakaryocytes \quad D. Platelets \\
\textbf{Answer:} D \\ 
\midrule

\textbf{Question 2} \\
Lindsay has been prescribed insulin therapy for which condition? \\
A. None of the other choices \quad B. Diabetes \quad \\
C. Hemophilia \quad D. Spina bifida \\
\textbf{Answer:} B \\ 
\midrule

\textbf{Question 3} \\
Why is it crucial to control and reduce the amount of dust that enters the air? \\
A. Less dust means less cleaning up afterwards \quad B. Dust in the air will affect your vision \\
C. Dust is always in the air and it does not cause harm \quad D. Constantly inhaling dust particles can cause lung problems in the future \\
\textbf{Answer:} D \\ 
\hline

\end{tabular}
\caption{Three Random Questions from the Business, Economics, and Health of the MMLU-CF Validation Set.}
\label{table_question3}
\end{table*}

\begin{table*}[h!]
\small
\centering
\begin{tabular}{|p{14cm}|}
\hline
\multicolumn{1}{|c|}{\textbf{History}} \\ \hline

\textbf{Question 1} \\
The constitutional history of France starts with the French Revolution in what year? \\
A. 1786 \quad B. 1780 \quad \\
C. 1789 \quad D. None of the other choices \\
\textbf{Answer:} C \\ \hline

\textbf{Question 2} \\
Between 1889 and 1916, where was the Second International, which developed under the influence of Socialist Philosophy, organized?\\
A. None of the other choices\\
B. London\\
C. Paris\\
D. Brussels\\
\textbf{Answer:} C \\ \hline

\textbf{Question 3} \\
What was the capital of the Hoysalas?\\
A. Dwarasamudra \quad B. Halebeedu\\
C. Sosevuru \quad D. Belur\\
\textbf{Answer:} A \\ \hline

\multicolumn{1}{|c|}{\textbf{Law}} \\ \hline

\textbf{Question 1} \\
How are computer programs legally safeguarded? \\
A. Copy rights. \quad B. Trademarks. \quad \\
C. Industrial design. \quad D. Patents. \\
\textbf{Answer:} A \\ \hline

\textbf{Question 2} \\
What type of justice is represented by the penalty imposed for breaking the law? \\
A. Political justice \quad B. Moral justice \quad \\
C. Legal justice \quad D. Economic justice \\
\textbf{Answer:} C \\ \hline

\textbf{Question 3} \\
What does WIPO stand for? \\
A. World Information and Patents Organisation \\
B. World Intellectual Property Organisation \\
C. World Information Protection Organisation \\
D. None of the other choices \\
\textbf{Answer:} B \\ \hline

\multicolumn{1}{|c|}{\textbf{Philosophy}} \\ \hline

\textbf{Question 1} \\
What does it mean when a reprehensible act is referred to by a different term? \\
A. None of the other choices \quad B. advantageous comparison \\
C. euphemistic labeling \quad D. attribution of blame \\
\textbf{Answer:} C \\ \hline

\textbf{Question 2} \\
The assertion, `Being non-violent is good' is a: \\
A. Religious judgement \quad B. None of the other choices \\
C. Factual judgement \quad D. Value judgement \\
\textbf{Answer:} D \\ \hline

\textbf{Question 3} \\
What does the phrase `lived alone on the forest tree' symbolize?\\
A. None of the other choices \quad B. Freedom \\
C. A dull life \quad D. A dependent life\\
\textbf{Answer:} B\\ \hline

\end{tabular}
\caption{Three Random Questions from the History, Law, and Philosophy of the MMLU-CF Validation Set.}
\label{table_question4}
\end{table*}

\begin{table*}[h!]
\small
\centering
\begin{tabular}{|p{14cm}|}
\hline
\multicolumn{1}{|c|}{\textbf{Psychology}} \\ \hline

\textbf{Question 1} \\
Which of the following happens first in development? \\
A. Secondary sexual characteristics \quad B. Reproductive maturity \\
C. Gender identity \quad D. Primary sexual characteristics \\
\textbf{Answer:} D \\ \hline

\textbf{Question 2} \\
How can a teacher be successful? \\
A. imparts subject knowledge to students \\
B. presents the subject matter in a well organized manner \\
C. prepares students to pass the examination \\
D. None of the other choices \\
\textbf{Answer:} B \\ \hline

\textbf{Question 3} \\
What is meant by Ex Post Facto research? \\
A. The research is carried out prior to the incident \\
B. None of the other choices \\
C. The research is carried out along with the happening of an incident \\
D. The research is carried out after the incident \\
\textbf{Answer:} D \\ \hline

\multicolumn{1}{|c|}{\textbf{Other}} \\ \hline

\textbf{Question 1} \\
To achieve a quick promotion, he came up with a plan to appease the manager. \\
A. Conciliate \quad B. Evict \quad C. Incite \quad D. Praise \\
\textbf{Answer:} A \\ \hline

\textbf{Question 2} \\
Which company initiated the secret Zuma Mission for the United States government? \\
A. SpaceX \quad B. None of the other choices \\
C. XCOR Aerospace \quad D. Boeing \\
\textbf{Answer:} A \\ \hline

\textbf{Question 3} \\
In The Calling of Saint Matthew, Caravaggio depicted his subjects wearing the clothing of his own era,\\
rather than that of Jesus's time. \\
A. to portray the painting’s patrons realistically. \\
B. to conform with other paintings in the series. \\
C. to enable the audience to identify with them. \\
D. so that he could use richer colors and brushstrokes. \\
\textbf{Answer:} C \\ \hline

\end{tabular}
\caption{Three Random Questions from the Psychology, Other of the MMLU-CF Validation Set.}
\label{table_question5}
\end{table*}

\end{document}